\documentclass[10pt,conference]{IEEEtran}
\IEEEoverridecommandlockouts
\usepackage{cite}
\usepackage{amsmath,amssymb,amsfonts}
\usepackage{algorithmic}
\usepackage{graphicx}
\usepackage{textcomp}
\usepackage{xcolor}
\usepackage{algorithm}
\usepackage{subcaption}
\usepackage{booktabs}
\usepackage[T1]{fontenc}
\usepackage{aecompl}

\def\x{{\mathbf x}}
\def\L{{\cal L}}
\newcommand{\R}{\mathbb{R}}  
\def\E{{\mathbb E}}
\def\w{{\mathbf w}}
\def\g{{\mathbf g}}
\def\m{{\mathbf m}}
\def\b{{\mathbf b}}
\def\x{{\mathbf x}}
\def\y{{\mathbf y}}
\def\D{{\mathcal D}}

\def\C{{\mathcal C}}
\newtheorem{theorem}{Theorem}
\newtheorem{lemma}{Lemma}
\newtheorem{corollary}{Corollary}
\newtheorem{assumption}{Assumption}
\def\BibTeX{{\rm B\kern-.05em{\sc i\kern-.025em b}\kern-.08em
    T\kern-.1667em\lower.7ex\hbox{E}\kern-.125emX}}

\begin{document}

\title{Adaptive Top-K in SGD for Communication-Efficient Distributed Learning\\}
\author{\IEEEauthorblockN{Mengzhe Ruan$^{1,2}$ \quad Guangfeng Yan$^{1,2}$\quad Yuanzhang Xiao$^{3}$ \quad Linqi Song$^{1,2}$\quad Weitao Xu$^{1,2,*}$\thanks{Weitao Xu is the corresponding author.}}
\IEEEauthorblockA{$^{1}$ City University of Hong Kong Shenzhen Research Institute, Shenzhen, China\\
$^{2}$ Department of Computer Science, City University of Hong Kong, Hong Kong SAR, China\\
$^{3}$Hawaii Advanced Wireless Technologies Institute, University of Hawaii at Manoa, Honolulu, HI, USA
}}

\maketitle

\begin{abstract}
Distributed stochastic gradient descent (SGD) with gradient compression has become a popular communication-efficient solution for accelerating distributed learning. One commonly used method for gradient compression is Top-K sparsification, which sparsifies the gradients by a fixed degree during model training. However, there has been a lack of an adaptive approach to adjust the sparsification degree to maximize the potential of the model's performance or training speed. This paper proposes a novel adaptive Top-K in SGD framework that enables an adaptive degree of sparsification for each gradient descent step to optimize the convergence performance by balancing the trade-off between communication cost and convergence error. Firstly, an upper bound of convergence error is derived for the adaptive sparsification scheme and the loss function. Secondly, an algorithm is designed to minimize the convergence error under the communication cost constraints. Finally, numerical results on the MNIST and CIFAR-10 datasets demonstrate that the proposed adaptive Top-K algorithm in SGD achieves a significantly better convergence rate compared to state-of-the-art methods, even after considering error compensation.
\end{abstract}

\section{Introduction}
\label{sec:intro}

Nowadays, with extensive data collected in distributed networks, there is an increasing need for distributed learning algorithms that aggregate local gradients to learn a global models. Distributed stochastic gradient descent (SGD) is the core of most distributed learning algorithms~\cite{dean2012large}. In practical networks, however, the communication overhead of transmitting gradients often becomes the performance bottleneck due to the limited bandwidth. Gradient compression, which uses less information to represent the gradients, is an effective and efficient method to solve this problem. The compression methods, however,  inevitably introduce compression noise which affects the convergence of the model. Therefore, how to choose the compression methods and the compression level efficiently to balance the trade-off between communication cost and convergence performance remains an open challenge.

Traditional compression methods often compress parameters with a \textit{fixed} compression factor for all the training iterations, which may not be optimal. To further improve communication efficiency, an online learning method was proposed in~\cite{han2020adaptive} to adaptively adjust the degree of gradient sparsity when the total dataset is non-i.i.d distributed in the federated learning network. Unfortunately, there lacks a theoretical convergence analysis in their research. In~\cite{yan2022ac}, an adaptive quantization method is proposed and its theoretical guarantee has also been proved. Nevertheless, the quantization method needs more computing resources than sparsification methods, which simply keep some components of the gradient and set others to zero.  
Therefore, we would like to investigate the adaptive sparsification methods in distributed SGD. We will improve upon Top-K, the most commonly-used biased sparsification method, which keeps only a few coordinates of the stochastic gradient with the largest magnitudes.


In this paper, we propose a novel adaptive Top-K SGD framework, named AdapTop-K, that aims to improve the convergence performance of Top-K while maintaining the same communication cost. 
Under the assumption of smoothness and Polyak-Lojasiewicz condition~\cite{karimi2016linear}, we derive an upper bound on the gap between the loss function and the optimal loss to characterize the convergence error caused by limited iteration steps, sampling, and adaptive Top-K sparsification. 
Based on the theoretical analysis, we design an adaptive Top-K method by minimizing the convergence upper bound under the desired total communication cost. The proposed AdapTop-K algorithm adjusts the degree of sparsification by considering the desired model performance, the number of rounds, and the norm of gradients. We validate our theoretical analysis through experiments on image classification tasks on the MNIST and CIFAR-10 datasets. Numerical results show that AdapTop-K outperforms the baseline sparsification methods.

To summarize, our key contributions are as follows:

$\bullet$ We propose a novel framework to characterize the trade-off between the communication cost and the convergence rate by adaptively adjusting the gradient sparsification levels in distributed learning. We analyze the convergence error of the loss function under Top-K sparsification for gradients over different communication rounds. We isolate our bound on the convergence error to characterize the impact of adaptive sparsification on the convergence rate.
    
$\bullet$ We solve the optimization problem that minimizes the convergence error while keeping the same communication cost as Top-K. To achieve this, we propose a novel adaptive Top-K algorithm called AdapTop-K, which dynamically adjusts the degree of gradient sparsification during training to improve model performance.

$\bullet$ We validate the proposed AdapTop-K on the popular datasets and machine learning models, demonstrating that our proposed AdapTop-K outperforms state-of-the-art gradient sparsification methods.

\section{Related Work}

There are two main approaches to compress SGD to reduce communication cost: quantization and sparsification.
Quantization compresses gradients by limiting the number of bits representing floating point numbers during communication. The gradient quantization was proposed in~\cite{alistarh2017qsgd}. There are several variants of quantization, including error compensation~\cite{wu2018error}, variance-reduced quantization~\cite{zhang2017zipml}, quantization to a ternary vector~\cite{wen2017terngrad}, and quantization of gradient difference~\cite{mishchenko2019distributed}. Sparsification methods aim to reduce the number of non-zero entries in the stochastic gradients~\cite{wangni2018gradient}. An aggressive sparsification method (Top-K)~\cite{stich2018sparsified} is to keep very few coordinates of the stochastic gradient with the largest magnitudes.
The methods can also be classified based on whether the compression is biased or unbiased. The unbiased methods could keep the expectation of compressed gradients as that of the true gradients~\cite{alistarh2017qsgd} and~\cite{wen2017terngrad}. In contrast, the biased methods introduce bias in the compression and more compression noise to the optimization process~\cite{stich2018sparsified}. These methods can compress the gradient efficiently to speed up distributed training. However, they do not consider adaptively changing the degree of compression during training, which is the key difference between our method and existing methods.


\section{System Model}

We consider a distributed learning system with a central server and $M$ edge devices (workers). The workers collaborate to train a shared machine learning model by aggregating the gradient or its variant in cooperation with the central server.

The learning model is represented by the vector of its parameters $\mathbf{w}\in \R^d$, where $d$ is the model size. The datasets are distributed over the $M$ workers. We use $\mathcal{D}^i$ to denote the local dataset at worker $i$. The global loss function, denoted by $F: \R^d \rightarrow \R$, is defined as
\begin{equation}
\begin{aligned}
F(\w)&=\frac{1}{M}\sum_{i=1}^{M}f^i(\w),\\ ~\text{with}~ f^i(\w)&=\E_{\xi^i\sim \D^i}\left[l^i(\w;\xi^i)\right],
\label{2.1}
\end{aligned}
\end{equation}
where $l^i(\w;\xi^i)$ is the local loss function of the model parameters $\w$ at work $i$, given the mini-batch $\xi^i$ randomly selected from worker $i$'s local dataset $\D^i$.

The objective of the training is to find a model parameter $\w$ to minimize the global loss function in Eq.~\eqref{2.1} :
\begin{equation}
\setlength{\abovedisplayskip}{-0.5pt}
\setlength{\belowdisplayskip}{-0.5pt}
\begin{aligned}
\w^\ast=\text{arg}\mathop{\text{min}}_{\w}F(\w).
\label{2.2}
\end{aligned}
\end{equation}

The distributed SGD is the most popular method to solve this problem, where each worker $i$ computes its local stochastic gradient $\g_t^i=\nabla l^i(\w_t;\xi^i)$ given parameters $\w_t$ at round $t$. Then the workers send the local gradient $\g_t^i$ to the central server. The server aggregates these gradients to update the model. To reduce the communication cost, we compress the local stochastic gradients before sending them to the server:
\begin{equation}
\setlength{\abovedisplayskip}{-1pt}
\setlength{\belowdisplayskip}{-1pt}
\begin{aligned}
\w_{t+1}=\w_t-\frac{\eta_t}{M}\sum_{i=1}^{M}\C^i[\g_t^i],
\label{2.3}
\end{aligned}
\end{equation}
where $\eta_t$ is the learning rate at iteration $t$, and $\C^i[\cdot]$ is the compression operator. Without the gradient compressor, Eq. \eqref{2.3} reduces to the vanilla distributed SGD with $\w_{t+1}=\w_t-\frac{\eta_t}{M}\sum_{i=1}^{M} \g_t^i$. The same procedure is repeated until the convergence criterion or the maximum number of communication rounds is reached.

A commonly-used compression operator is Top-K, where each worker $i$ keeps only $k$ elements of the gradient $\g_t^i$ with the largest magnitudes and sets the other elements to zero \cite{stich2018sparsified}. In this work, we speed up the convergence of Top-K by adaptively choosing the sparsity of the gradient during the convergence process. Specifically, given a total of $T$ rounds of gradient update, our goal is to find the optimal sparsity levels $k_0, \ldots, k_{T-1}$ in each round, so that the final model is as close to the optimal model as possible. It is natural to measure the gap from the optimal model by the difference between the expectation of the final global loss $F(\w_T)$ and the optimal loss $F^{\ast}=F(\w^{\ast})$. Note that we need to take expectation of the final loss $F(\w_T)$ due to the stochastic gradient descent. Therefore, our design problem can be formulated as follows
\begin{eqnarray}\label{2.8}
\setlength{\abovedisplayskip}{-2pt}
\setlength{\belowdisplayskip}{-2pt}
& \mathop{\text{min}} \limits_{k_0,\ldots, k_{T-1}} & \E\left[F(\w_T)\right]-F^*\\
& \text{s.t.} & \textstyle\sum_{t=0}^{T-1} k_t\leq K, \nonumber \\
&             & k_t \in \{0, 1, \ldots, d\}, ~ t=0,\ldots,T-1, \nonumber
\end{eqnarray}
where $K$ is the total budget for the communication overhead during the training. When comparing with other sparsification methods, we can set the communication budget $K$ accordingly.

\section{Proposed Algorithm}

In this section, we first provide convergence analysis of AdapTop-K given a sequence of sparsity levels $k_0, \ldots, k_{T-1}$. Based on the analysis, we then propose a practical algorithm for finding a sequence $k_0, \ldots, k_{T-1}$ that guarantees to outperform the standard Top-K method.

\subsection{Convergence Analysis}

For the convergence analysis, we make standard assumptions on the stochastic gradient and the loss function that are commonly used in the literature~\cite{yan2022ac},~\cite{Li2020On}, and~\cite{cao2021optimized}.
\begin{assumption}\label{ass:1}(Smoothness). There exists a non-negative constant $L$ such that for any $\x,\y\in\R^d$,
\end{assumption}
\begin{equation}
\setlength{\abovedisplayskip}{-1pt}
\setlength{\belowdisplayskip}{-1pt}
\begin{aligned}
F(\x)-F(\y)-\langle \nabla F(\y),\x-\y\rangle\leq\frac L2\Vert \x-\y\Vert^2,
\label{2.4}
\end{aligned}
\end{equation}
where $\nabla F(\y)$ is the gradient of the loss function $F(\cdot)$ at $\y$. 

\begin{assumption}\label{ass:2} (Polyak-Lojasiewicz Condition). There exists a constant $\mu\ge 0$ such that for any $\w\in\R^d$, we have
\end{assumption}
\begin{equation}
\begin{aligned}
\setlength{\abovedisplayskip}{-0pt}
\setlength{\belowdisplayskip}{-1pt}
\Vert \nabla F(\w)\Vert^2\geq 2\mu(F(\w)-F^\ast).
\label{2.5}
\end{aligned}
\end{equation}

Note that Assumption~\ref{ass:2} is milder than the assumption of strong convexity~\cite{karimi2016linear}.

\begin{assumption}\label{ass:3}(Unbiasedness and Bounded Variance of Stochastic
Gradient). The local stochastic gradients $\g^i$ are assumed to be independent and unbiased estimates of the local gradient $\nabla f^i(\w_t)$ with bounded variance:
\end{assumption}
\begin{eqnarray}
\E_{\xi^i \sim \D^i} \left[\g_t^i\right] &=& \nabla f^i(\w_t), \\
\E_{\xi^i \sim \D^i} \left[\Vert \g_t^i-\nabla f^i(\w_t)\Vert^2 \right] &\le& \sigma^2. \nonumber
\label{2.6}
\end{eqnarray}

As proven in~\cite{ajalloeian2020convergence}, the gradient update in Eq.~\eqref{2.3} can be rewritten as
\begin{equation}
\begin{aligned}
\w_{t+1}=\w_t-\eta_t \C[\g_t], \label{3.1}
\end{aligned}
\end{equation}
where $\g_t$ is the stochastic gradient of the global loss function
\begin{equation}\label{eqn:stochastic_gradient_global}
\g_t=\nabla F(\w_t)+\m_t(\w_t),
\end{equation}
and $\C[\cdot]$ is the aggregate Top-K operator
\begin{equation}
\C[\g_t]=\nabla F(\w_t)+\m_t(\w_t)+\b_t(\w_t),
\label{3.2}
\end{equation}
where $\m_t(\w_t)$ is the noise in SGD and $\b_t(\w_t)$ is the bias introduced by sparsification. 

By Assumption~\ref{ass:3}, the noise has zero mean and bounded variance, namely
\begin{equation}
\begin{aligned}
\E[\m_t(\w_t)] = 0\quad \text{and} \quad \E[\Vert \m_t(\w_t)\Vert^2]\leq \sigma^2.
\label{3.3}
\end{aligned}
\end{equation}

An upper bound of the variance of the bias is given in \cite{stich2018sparsified}. We summarize the result as a lemma here.
\begin{lemma}\label{lemma:1}(Bounded Variance of Stochastic Gradient with Top-K sparsification). The variance of the bias $\b_t(\w_t)$ is upper bounded by the mini-batch gradient $\g_t$ as follows:~\cite{stich2018sparsified}
\end{lemma}
\begin{equation}
\Vert \b_t(\w_t)\Vert^2\leq \left(1-\frac{k}{d}\right)\Vert \g_t \Vert^2.
\label{3.4}
\end{equation}

With Lemma~\ref{lemma:1}, we prove an upper bound of the optimality gap under the adaptive sparsity levels of $k_0,\ldots,k_{T-1}$.

\begin{theorem}\label{theo:1} (Upper Bound for Convergence Error). Under Assumptions~\ref{ass:1}--\ref{ass:3}, given the initial parameter $\w_0$ and constant stepsize $\eta_t=\eta\leq\frac{1}{L}$, the optimality gap of the adaptive Top-K method is upper bounded by
\begin{eqnarray}\label{3.5}
\E[F(\w_{T})]-F^* \!\!\!\!&\leq& \!\!\!\!\underbrace{\left(1-\frac{\eta \mu }{d}k\right)^T \left[F(\w_0)-F^*\right]}_{\mathbf{M}(k)}\\
&+&\!\!\!\!\underbrace{\frac{d \sigma^2}{2k\mu}\left(1-\frac{k}{d}+\eta L\right)\left[1-\left(1-\frac{\eta \mu }{d}k\right)^{T}\right]}_{\mathbf{N}(k)} \nonumber\\
&-&\!\!\!\!\underbrace{\sum_{t=0}^{T-1}\left[\left(\frac{\eta n_t}{2d}\Vert \g_t\Vert^2\right)\left(1-\frac{\eta \mu }{d}k\right)^{T-1-t}\right]}_{\text{the only term that depends on $n_t$}}, \nonumber
\end{eqnarray}
where $k = \frac{K}{T}$ is the average sparsity level and $n_t = k_t - k$ is the deviation from the average sparsity level at round $t$.
\end{theorem} 
\begin{IEEEproof}
See the appendix. The proofs of all the other results can be found in our technical report~\cite{2210.13532}.
\end{IEEEproof}

The upper bound in \eqref{3.5} has two parts. The first part is the sum of the first two terms $\mathbf{M}(k)+ \mathbf{N}(k)$, which depends only on the average sparsity level $k$. The second part is the third term, which is the only term that depends on $n_0, \ldots, n_{T-1}$. When $n_t=0$ for all $t$, the upper bound reduces to $\mathbf{M}(k)+ \mathbf{N}(k)$, namely the bound for the vanilla Top-K method.

\subsection{The Proposed AdapTop-K Algorithm}

We aim to minimize the upper bound of the optimality gap in \eqref{3.5} by choosing $n_0,\ldots,n_{T-1}$. Since only the third term depends on the adjustments $n_0,\ldots,n_{T-1}$, the optimization problem can be formulated as
\begin{eqnarray}\label{eqn:optimization_problem}
& \mathop{\text{max}}\limits_{n_0,\ldots,n_{T-1}} & \sum_{t=0}^{T-1}\left[\left(\frac{\eta n_t}{2d}\Vert \g_t\Vert^2\right)\left(1-\frac{\eta \mu }{d}k\right)^{T-1-t}\right]\\
& \text{s.t.} & \sum_{t=0}^{T-1} n_t \le 0,  \nonumber \\
& & n_t \in \left\{-k, \ldots, d-k\right\}, ~ t=0,\ldots,T-1, \nonumber
\end{eqnarray}
where the first constraint comes from the constraint on the communication overhead in \eqref{2.8} and the second constraint comes from the fact that $k_t \in \{0, \ldots, d\}$.

Since the objective function is linear in $n_t$, the optimal solution should assign the largest possible values to the $n_t$'s with the largest coefficients
\begin{eqnarray}\label{eqn:coefficient_in_objective_function}
    \left(\frac{\eta}{2d}\Vert \g_t\Vert^2\right)\left(1-\frac{\eta \mu }{d}k\right)^{T-1-t},
\end{eqnarray}
subject to the upper bound $d-k$ and the budget of total communication overhead. However, the major challenge is that the coefficients in \eqref{eqn:coefficient_in_objective_function} depend on the gradients $\g_t$, which are stochastic due to the randomly selected mini-batches and are dependent on our choice of sparsity levels $n_0,\ldots,n_{t-1}$ up to round $t$. Therefore, we cannot solve the optimization problem \eqref{eqn:optimization_problem} directly. Instead, we choose to maximize an upper bound of the objective function, which is obtained by bounding the norm of the stochastic gradients $\g_t$.

\begin{lemma}\label{lemma:2} (Upper Bound for Stochastic Gradient). Under Assumptions~\ref{ass:1}--\ref{ass:3}, given the initial parameter $\w_0$ and constant stepsize $\eta_t=\eta\leq\frac{1}{L}$, the stochastic gradient in Eq.~\eqref{eqn:stochastic_gradient_global} can be upper bounded by
\end{lemma}
\begin{equation}
\begin{aligned}
\E[\|\g_t\|^2] \leq \frac{2d}{k\eta}\cdot\frac{F(\w_0)}{t }+\frac{d\sigma^2 }{k}(\eta L+1) \triangleq\frac{\alpha}{t}+\beta
\label{3.7}
\end{aligned}
\end{equation}
where $\alpha \triangleq \frac{2d}{k\eta}F(\w_0)$ and $\beta \triangleq \frac{d\sigma^2 }{k}(\eta L+1)$.

Based on Lemma~2, we obtain the following upper bound of the objective function in \eqref{eqn:optimization_problem}
\begin{eqnarray}\label{eqn:upper_bound_objective_function}
& & \frac{\eta}{2d} \sum_{t=0}^{T-1}\left[\left(\frac{\alpha}{t} + \beta\right)\left(1-\frac{\eta \mu }{d}k\right)^{T-1-t}\right] \cdot n_t \nonumber \\
&\triangleq& \frac{\eta}{2d} \sum_{t=0}^{T-1} (A_t B_t) \cdot  n_t,
\end{eqnarray}
where $A_t \triangleq \frac{\alpha}{t} + \beta$ and $B_t\triangleq (1-\frac{\eta \mu }{d}k)^{T-1-t}$. 

Finally, the optimization problem to solve is
\begin{eqnarray}\label{eqn:optimization_problem_upper_bound}
& \mathop{\text{max}}\limits_{n_0,\ldots,n_{T-1}} & \frac{\eta}{2d} \sum_{t=0}^{T-1} (A_t B_t) \cdot  n_t\\
& \text{s.t.} & \sum_{t=0}^{T-1} n_t \le 0,  \nonumber \\
& & n_t \in \left\{-k, \ldots, d-k\right\}, ~ t=0,\ldots,T-1. \nonumber
\end{eqnarray}

The objective function in \eqref{eqn:optimization_problem_upper_bound} is linear in $n_t$ with coefficient $A_t B_t$. We can prove the following monotonicity results.

\begin{lemma}\label{lemma:monotonicity}
    The coefficient $A_t B_t$ first decreases with $t$ and then increases with $t$. Specifically, we have
    \begin{eqnarray}
        & & A_{t+1} B_{t+1} < A_t B_t, ~\text{for}~t < \hat{t} \triangleq \left\lfloor\frac{-\alpha+\sqrt{\Delta }}{2\beta}\right\rfloor, ~\text{and} \nonumber \\
        & & A_{t+1} B_{t+1} \geq A_t B_t, ~\text{for}~t \geq \hat{t},
    \label{3.8}
    \end{eqnarray}
    where $\Delta \triangleq \alpha^2-\frac{4\alpha\beta}{lnB}$, $B \triangleq 1-\frac{\eta \mu }{d}k$, and $\lfloor\cdot\rfloor$ is the floor function.

\end{lemma}

Given the monotonicity result in Lemma~\ref{lemma:monotonicity}, we design the following adaptive sparsity levels
\begin{equation}
\setlength{\belowdisplayskip}{1pt}
\begin{aligned}
\left\{
    \begin{array}{cc}
        n_t=+\gamma k\Rightarrow{}k_t=(1+\gamma) k, & t\in[0,\frac{\hat{t}}{2})\cup [\frac{\hat{t}+T}{2},T-1]\\
        n_t=-\gamma k\Rightarrow{}k_t=(1-\gamma) k,& t\in [\frac{\hat{t}}{2},\frac{\hat{t}+T}{2}),
    \end{array}
\right.
\label{3.9}
\end{aligned}
\end{equation}
where $\gamma$ is the scaling factor (i.e., a hyperparameter). In the above scheme, $n_t$ takes the negative value half the training time and the positive value the other half, which satisfies the communication budget constraint. To maximize the objective function, we set $n_t$ to be positive when  $A_tB_t$ is larger.

\addtolength{\topmargin}{0.051in}

We can prove that the above adaptive sparsity levels result in a lower convergence error compared to the vanilla Top-K.

\begin{corollary}\label{coro:1}(Convergence Error Bound using AdapTop-K in distributed SGD). Under the adaptive sparsity levels in Eq.~\eqref{3.9}, the optimality gap is upper bounded by
\begin{equation}
\begin{aligned}
\E[F(\w_{T})]-F^*&\leq \mathbf{M}(k)+\mathbf{N}(k)\\
+\frac{\eta \gamma k}{2d}&\underbrace{\left(\sum_{t=\frac{\hat{t}}{2}}^{\frac{\hat{t}+T-1}{2}}A_tB_t-\sum_{t=0}^{\frac{\hat{t}}{2}}A_tB_t-\sum_{t=\frac{\hat{t}+T-1}{2}}^{T-1}A_tB_t\right)}_{\text{always less than 0 because of (\ref{3.8})}}\\
&\textless\quad \underbrace{\mathbf{M}(k)+\mathbf{N}(k)}_{\text{upper bound for SGD with vanilla Top-K}}.
\label{3.10}
\end{aligned}
\end{equation}
\end{corollary}

The pseudo-code of distributed SGD with the proposed AdapTop-K method is provided in Algorithm~\ref{algo:1}.

\begin{algorithm}
	\renewcommand{\algorithmicrequire}{\textbf{Input:}}	\renewcommand{\algorithmicensure}{\textbf{Output:}}
	\caption{AdapTop-K in Distributed SGD}
    \label{algo:1}
	\begin{algorithmic}[1]
        \REQUIRE Maximum iterations number $T$, learning rate $\eta$, initial point $\w_0\in\R^d$, fixed $k$ value, adjusted scale factor $\gamma$, hyper-parameters $\hat{t}$
        \ENSURE  $\w_t$
		\FOR {$t=0,1,...T-1$ }
		\STATE \textbf{On each worker} $i=1,...,M$:
		\STATE Compute stochastic local gradient $\g_t^i$
        \IF{$t\in [\frac{\hat{t}}{2},\frac{\hat{t}+T}{2})$}
        \STATE Set $k_t$ to $k-\gamma k$
        \ELSE
        \STATE Set $k_t$ to $k+\gamma k$
        \ENDIF
        \STATE Compress gradient $\g_t^i$ to $\C_{k_t}[\g_t^i]$
        \STATE Send $\C_{k_t}[\g_t^i]$ to server
		\STATE Receive $\w_{t+1}$ from server
        \STATE \textbf{On server}:
        \STATE Collect $M$ compressed gradients $\C_{k_t}[\g_t^i]$ from workers
        \STATE Aggregation: $\C_{k_t}[\g_t]=\sum_{i=1}^M\C_{k_t}[\g_t^i]$
        \STATE Update global parameters: $\w_{t+1}=\w_{t}-\frac{\eta}{M}\C_{k_t}[\g_t] $
		\STATE Send $\w_{t+1}$ back to all workers
		\ENDFOR  
	\end{algorithmic}  
\end{algorithm}

\begin{figure*}[htbp]
\begin{minipage}[t]{0.32\linewidth}
  \centering
  \includegraphics[width=4.8cm]{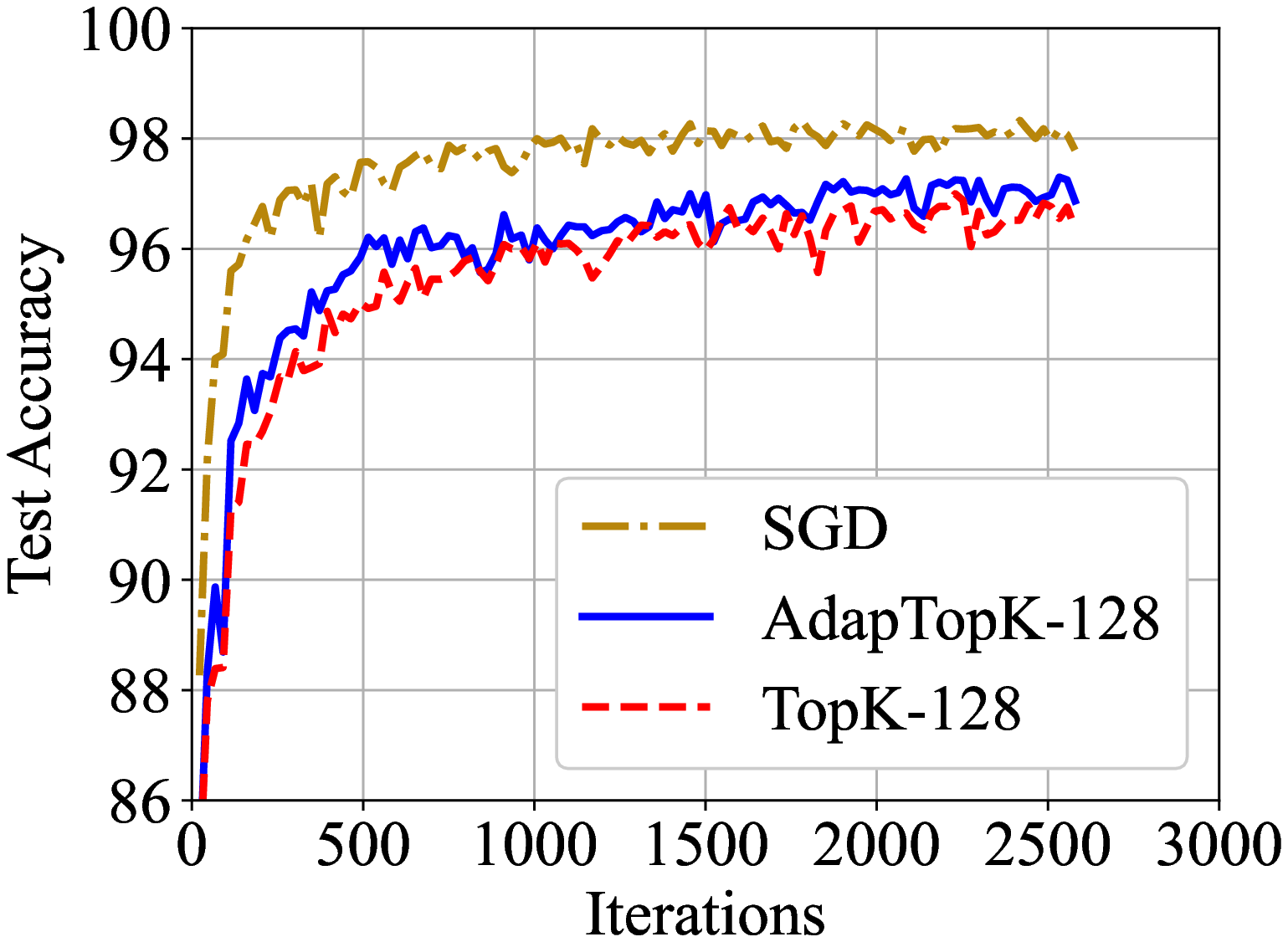}
  \subcaption{Accuracy ($\frac{d}{k}$=128)}
  \label{res:a}
\end{minipage}
\hfill
\begin{minipage}[t]{0.32\linewidth}
  \centering
  \centerline{\includegraphics[width=4.8cm]{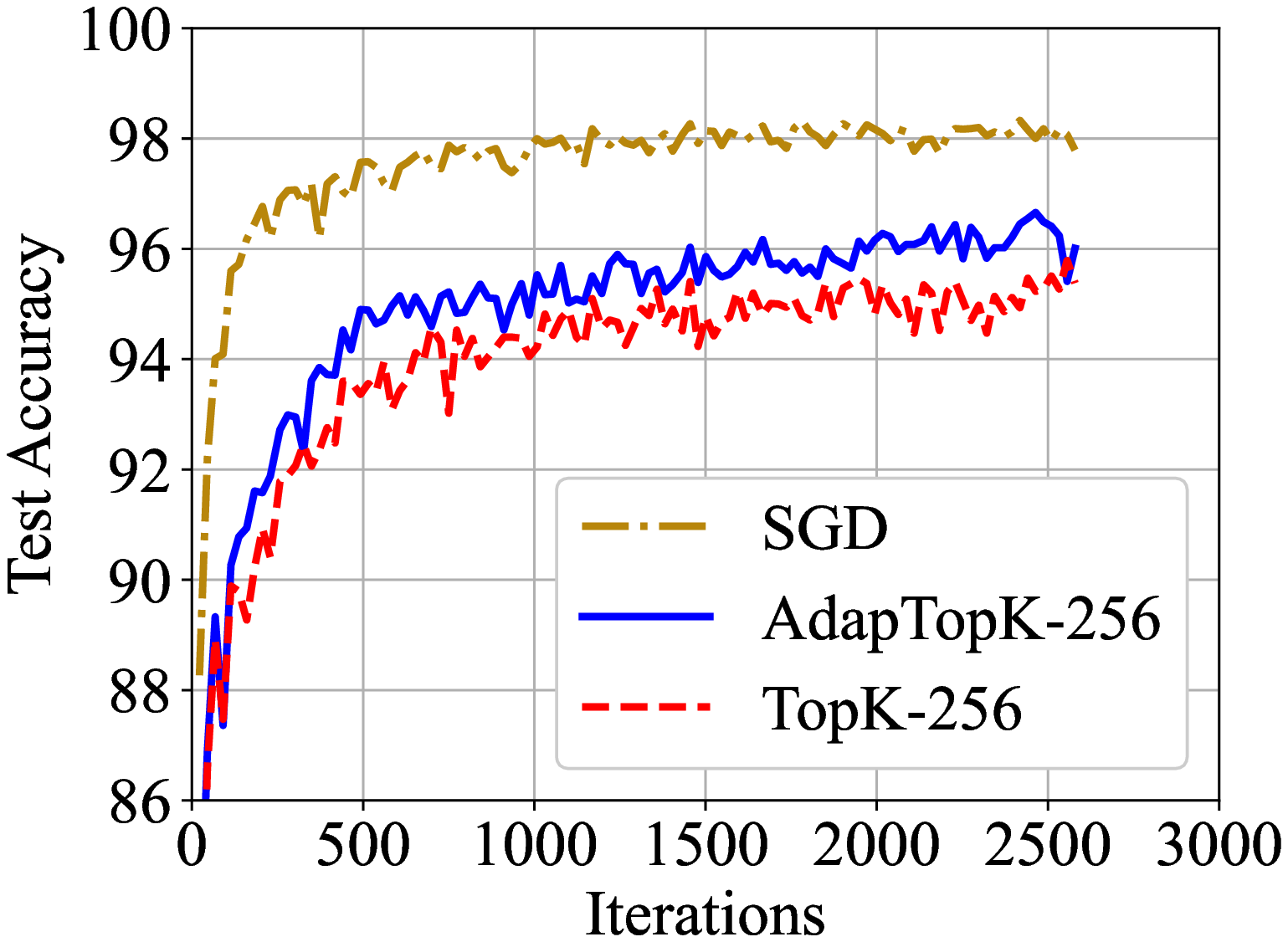}}
  \subcaption{Accuracy ($\frac{d}{k} $=256)}
  \label{res:b}
\end{minipage}
\hfill
\begin{minipage}[t]{0.32\linewidth}
  \centering
  \centerline{\includegraphics[width=4.8cm]{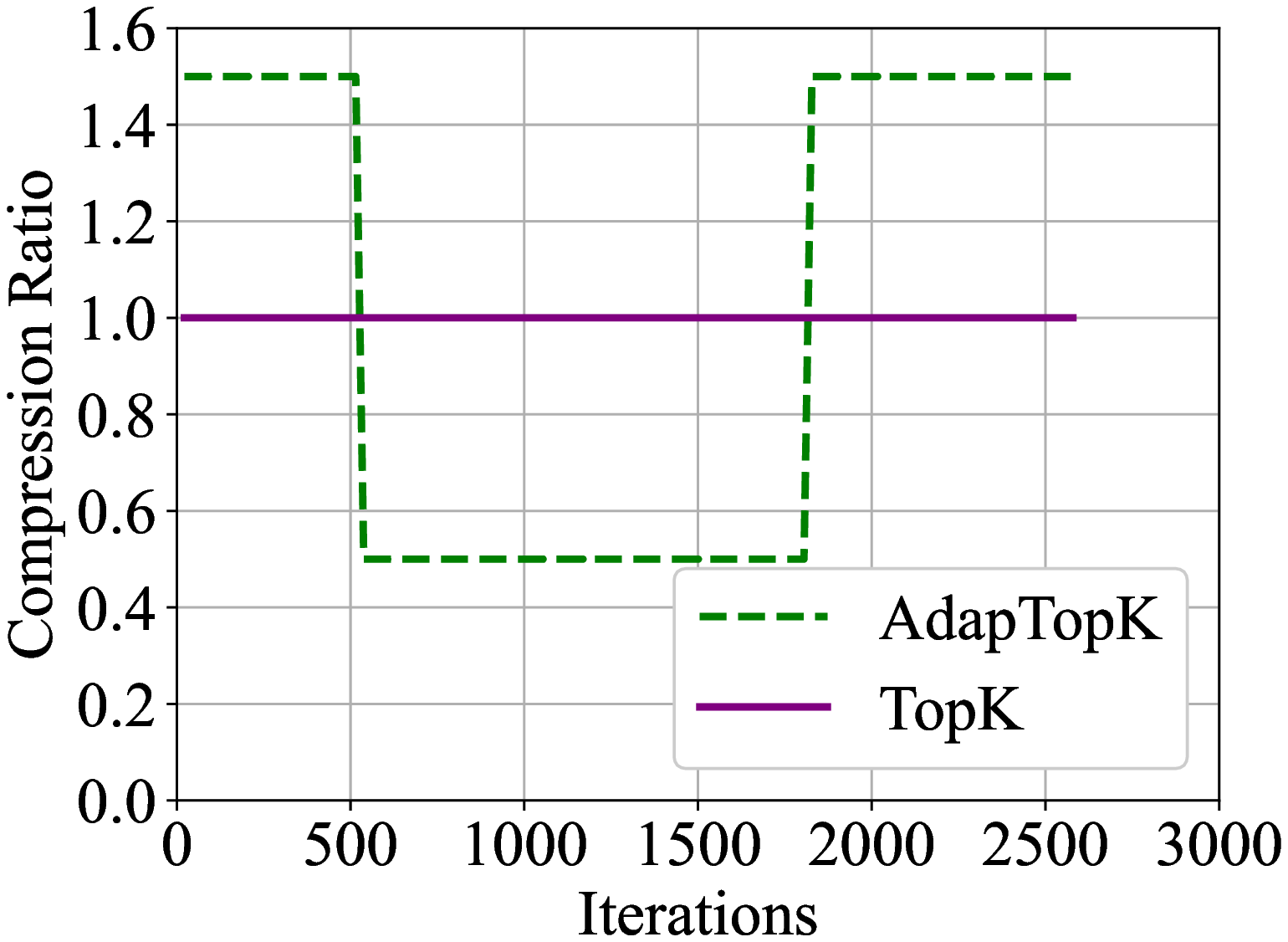}}
  \subcaption{Compression level on MNIST Dataset}
  \label{res:c}
\end{minipage}
\caption{Evaluation results of different methods on MNIST Dataset.}
\label{fig:res1}
\end{figure*}
\begin{figure*}[htbp]
\begin{minipage}[b]{0.32\linewidth}
  \centering
  \centerline{\includegraphics[width=4.8cm]{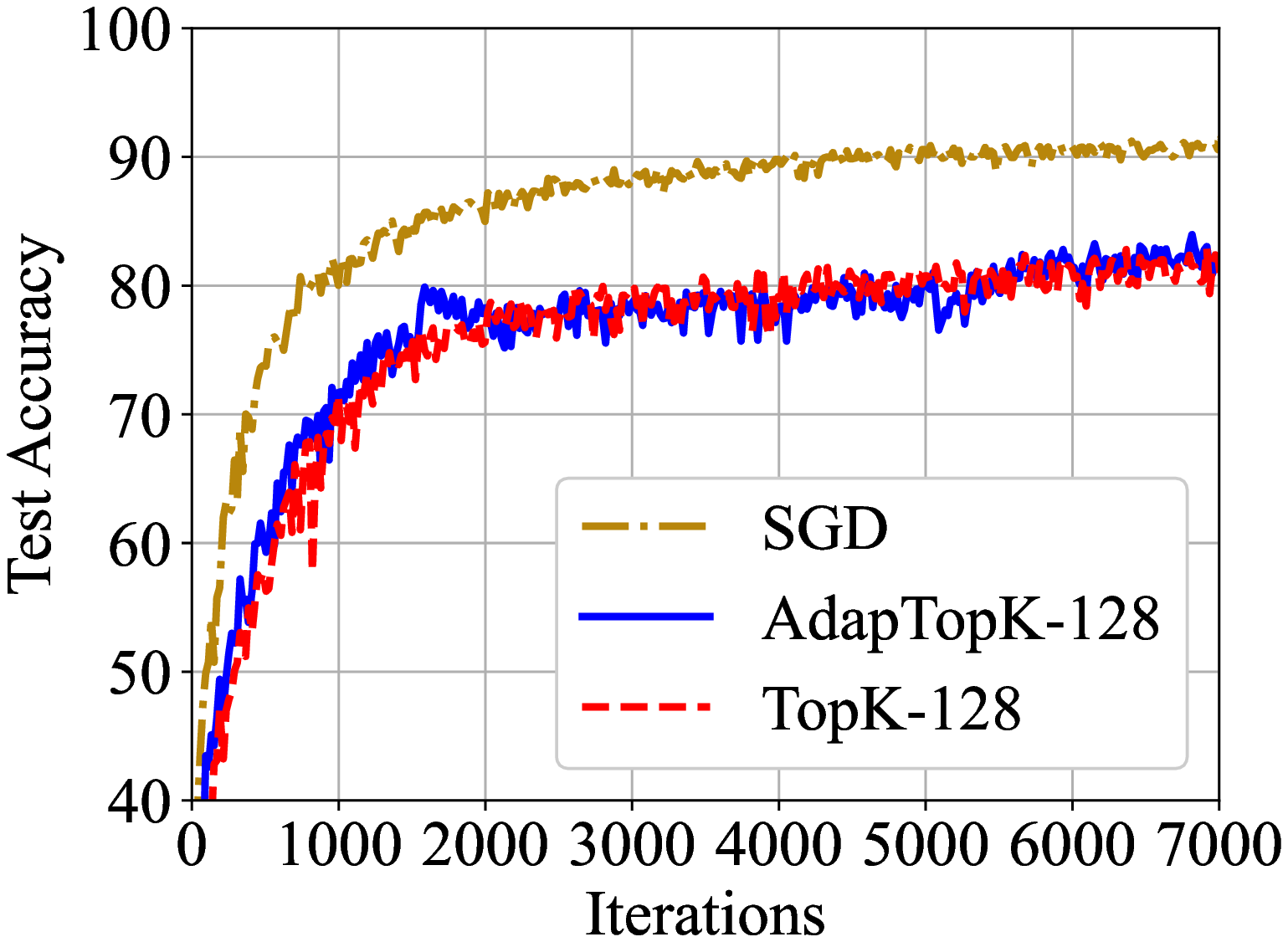}}
  \subcaption{Accuracy ($\frac{d}{k} $=128)}
  \label{res:d}
\end{minipage}
\hfill
\begin{minipage}[b]{0.32\linewidth}
  \centering
  \centerline{\includegraphics[width=4.8cm]{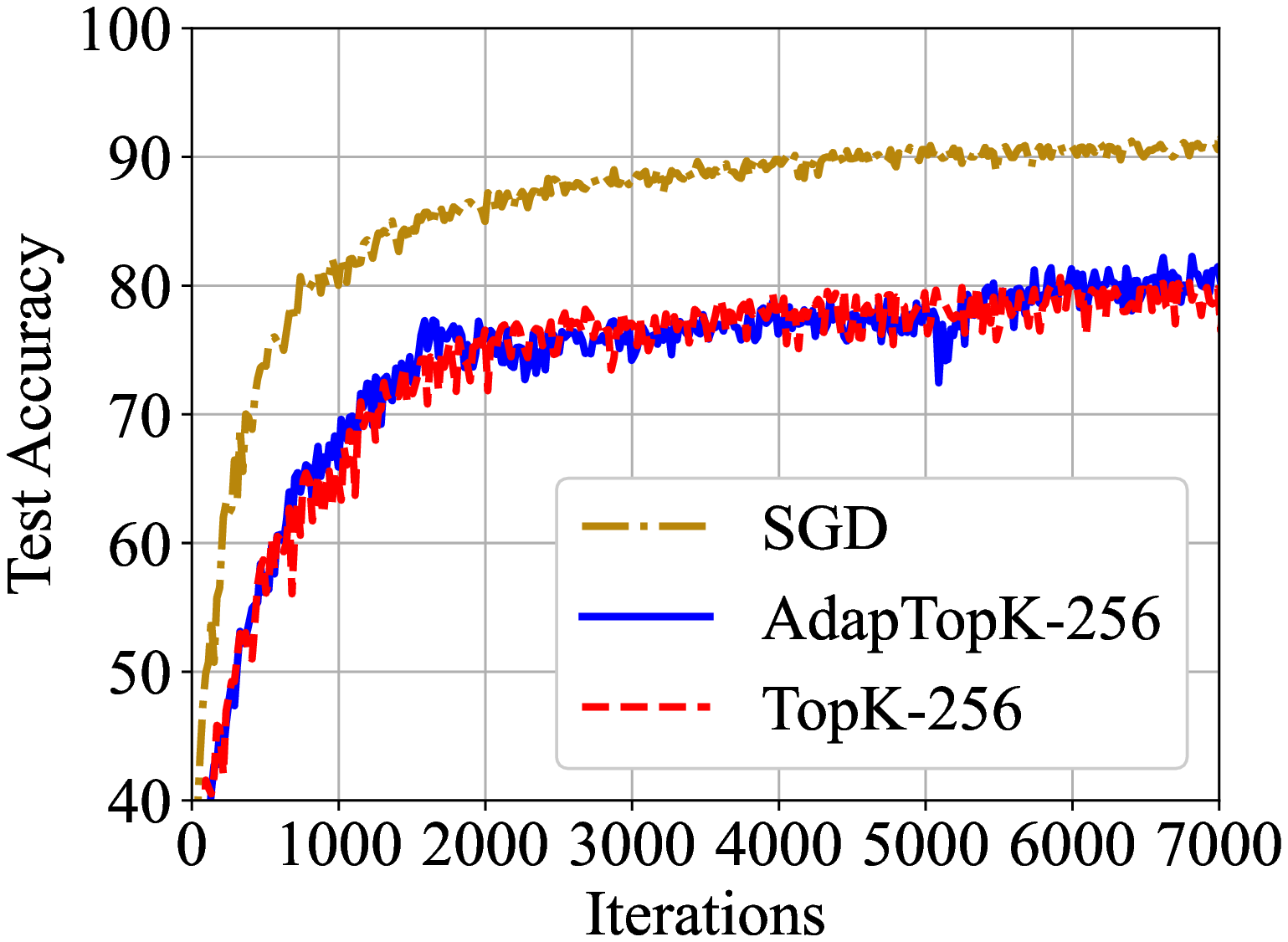}}
  \subcaption{Accuracy ($\frac{d}{k} $=256)}
  \label{res:e}
\end{minipage}
\hfill
\begin{minipage}[b]{0.32\linewidth}
  \centering
  \centerline{\includegraphics[width=4.8cm]{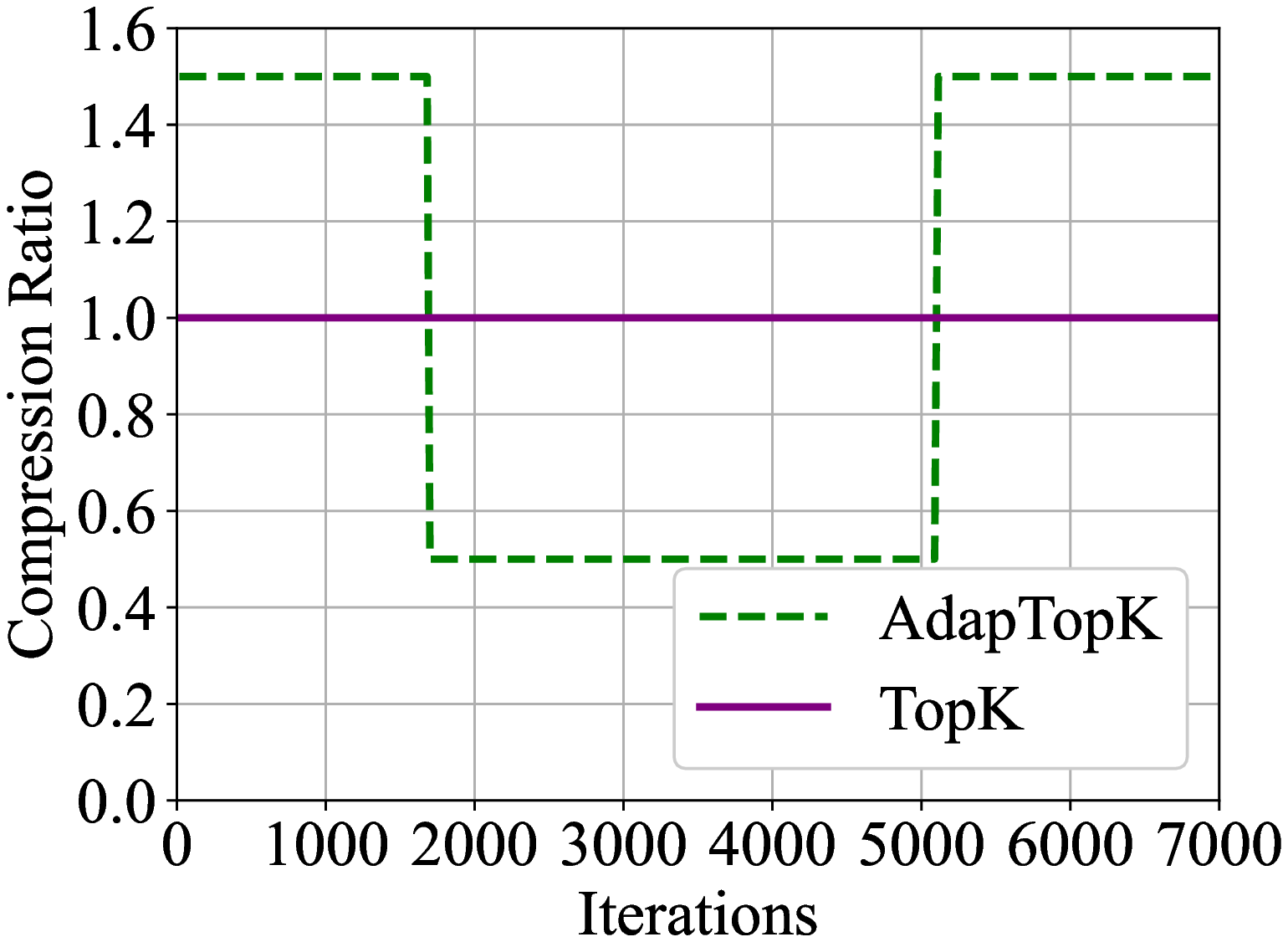}}
  \subcaption{Compression level on CIFAR-10 Dataset}
  \label{res:f}
\end{minipage}
\caption{Evaluation results of different methods on CIFAR-10.}
\label{fig:res2}
\end{figure*}

\begin{figure}[htb]
\begin{minipage}[b]{0.48\linewidth}
  \centering
  \centerline{\includegraphics[width=4.2cm]{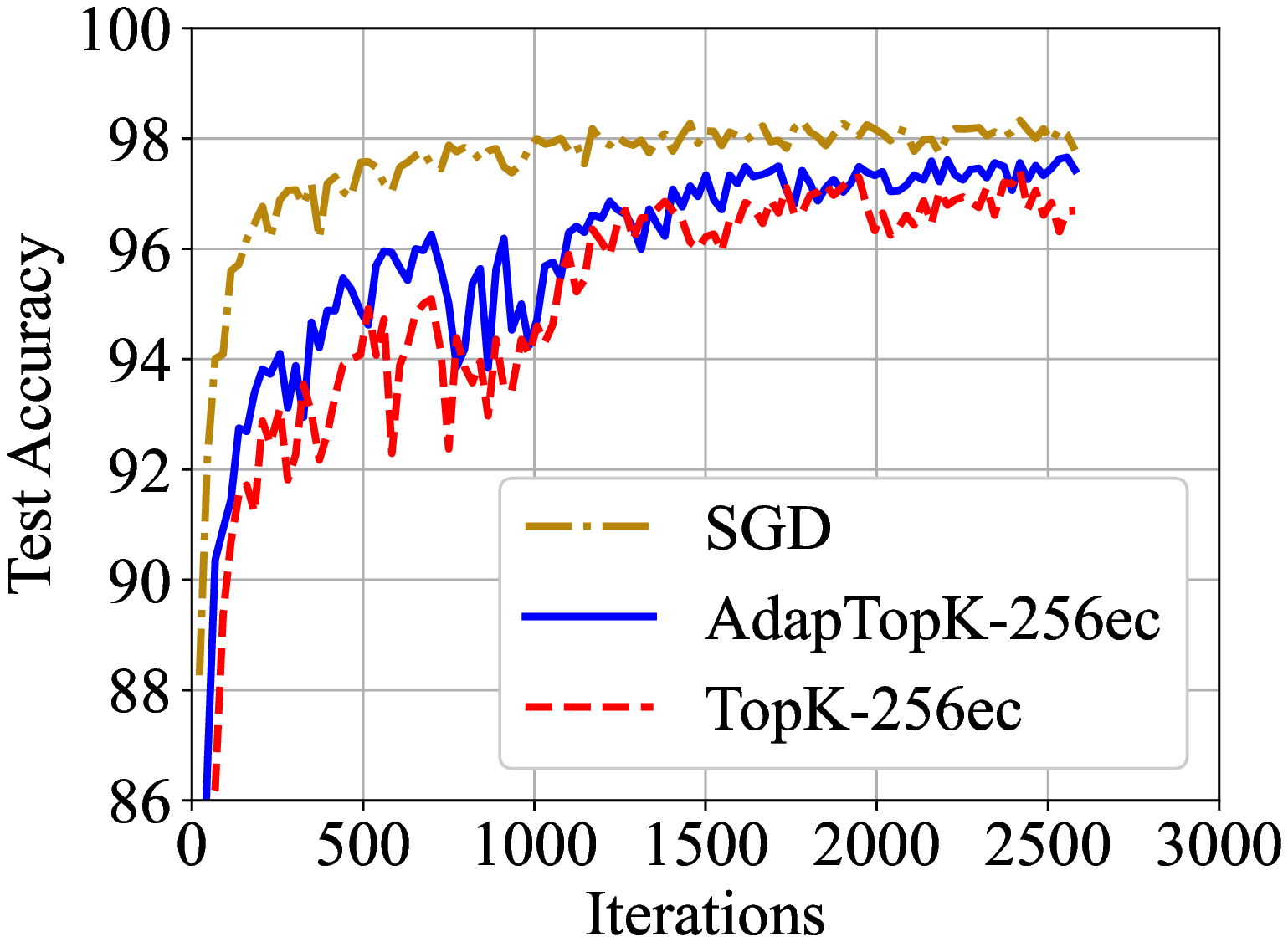}}
  \subcaption{Accuracy with ec ($\frac{d}{k} $=256)}
  \label{res3:a}
\end{minipage}
\hfill
\begin{minipage}[b]{0.48\linewidth}
  \centering
  \centerline{\includegraphics[width=4.2cm]{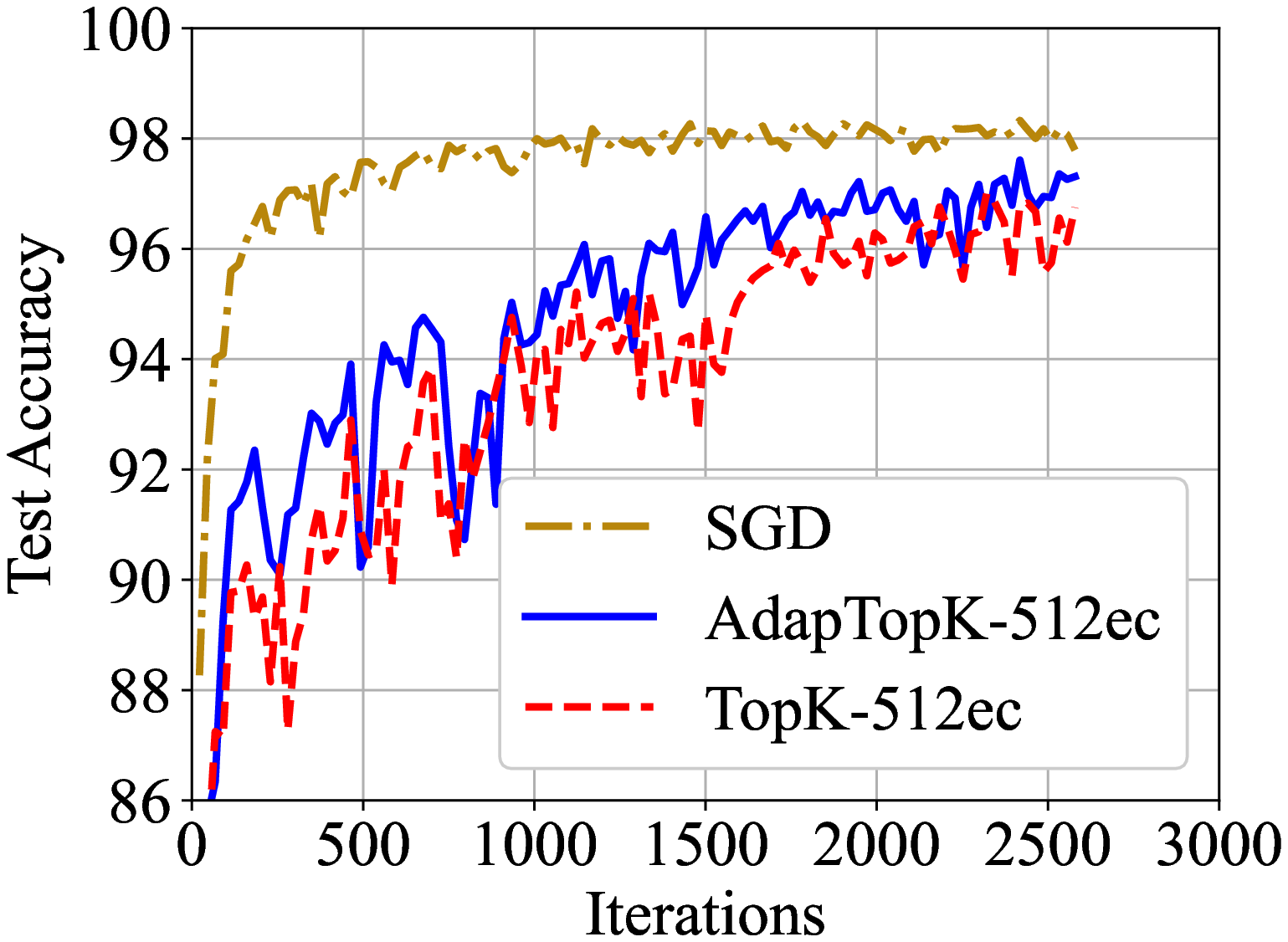}}
  \subcaption{Accuracy with ec ($\frac{d}{k} $=512)}
  \label{res3:b}
\end{minipage}
\caption{Evaluation with error compensation on MNIST.}
\label{fig:res3}
\end{figure}

\begin{figure}[htb]
\begin{minipage}[b]{0.48\linewidth}
  \centering
  \centerline{\includegraphics[width=4.2cm]{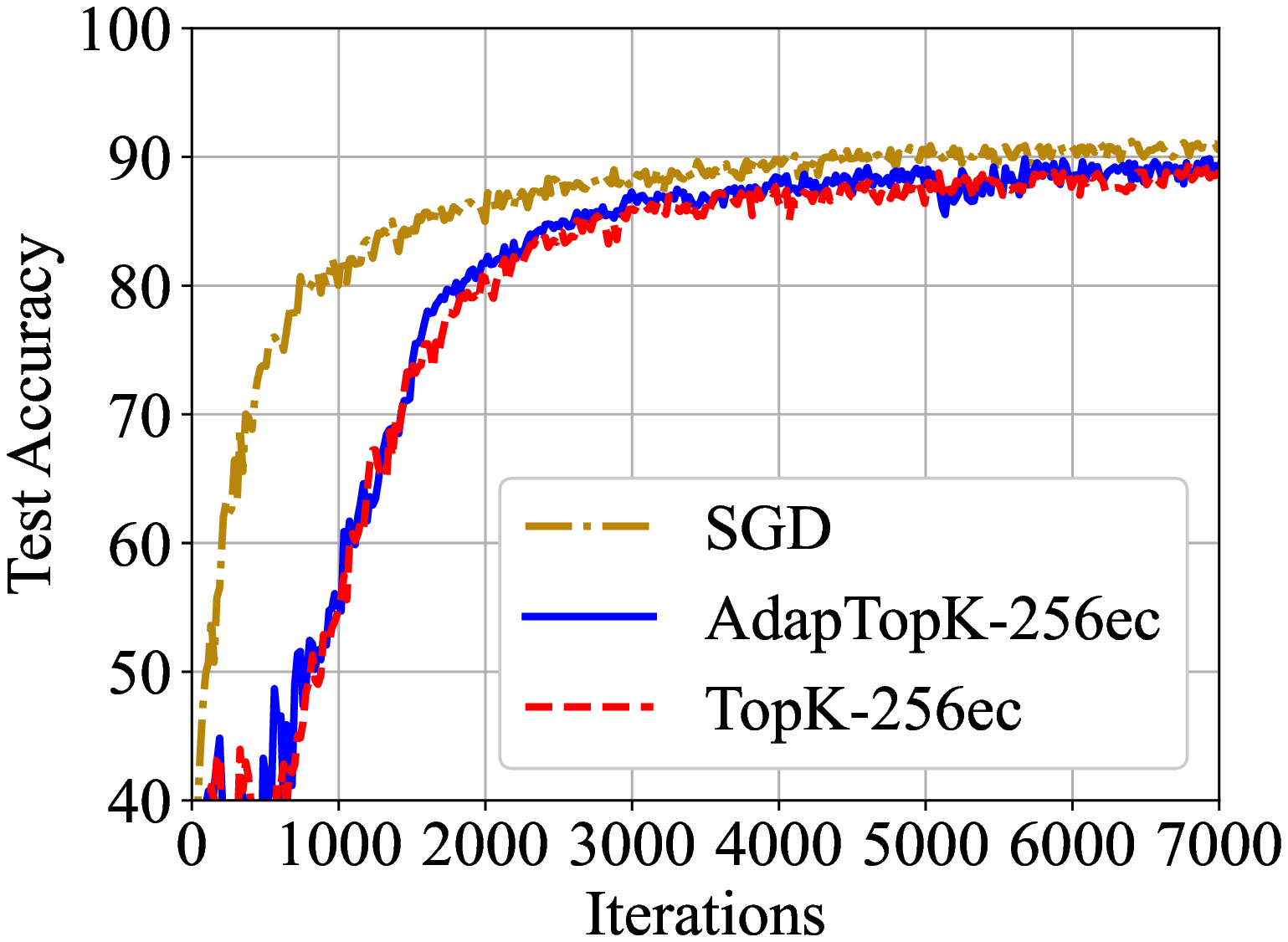}}
  \subcaption{Accuracy with ec ($\frac{d}{k} $=256)}
  \label{res4:a}
\end{minipage}
\hfill
\begin{minipage}[b]{0.48\linewidth}
  \centering
  \centerline{\includegraphics[width=4.2cm]{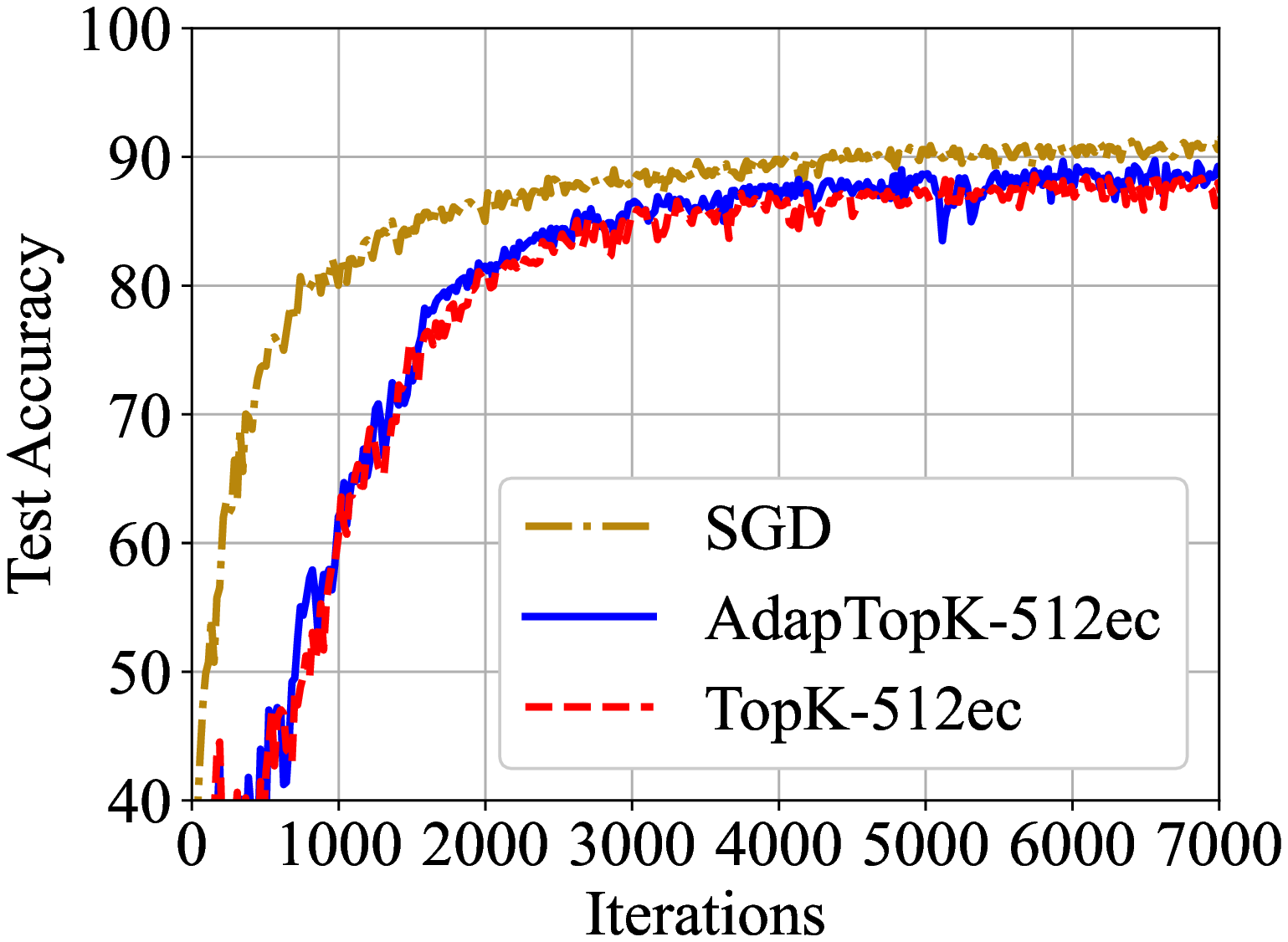}}
  \subcaption{Accuracy with ec ($\frac{d}{k} $=512)}
  \label{res4:b}
\end{minipage}
\caption{Evaluation with error compensation on CIFAR-10.}
\label{fig:res4}
\end{figure}


\section{Evaluation}
\vspace{-0.4mm}
\label{sec:illust}
In this section, we conduct experiments on two widely used datasets, namely MNIST and CIFAR-10, to validate the effectiveness of our proposed AdapTop-K method. We conduct experiments for $M$ = 8 workers and use canonical networks to evaluate the performance on the image classification task using different algorithms: fully-connected network on the MNIST dataset, and Resnet18 on the CIFAR-10 dataset. The above datasets are the database commonly used for training various image processing systems. Other parameters information is shown in Table~\ref{tab:1}. We use test accuracy to measure the learning performance. We compare our proposed AdapTop-K in SGD with the vanilla Top-K. 
\begin{table}[htb]
\vspace{-0.5em}
    \centering
    \begin{tabular}{c|c|c}
        \toprule
        Dataset & MNIST & CIFAR-10 \\
        \midrule
        Networks & fully-connected network & ResNet18 \\
        Model Size & $d=785$ & $d=1\times 10^7$ \\
        Learning Rate & 0.1 & 0.1 \\
        Batch Size & 32 & 32 \\
        Workers & 8 & 8 \\
        Iterations & 3,000 & 7,000 \\
        Compression Ratio &  128/256/512 &  128/256/512 \\
        $\gamma$ & 0.5 & 0.5\\
        \bottomrule
    \end{tabular}
    \caption{Experimental Setting.}
    \label{tab:1}
    \vspace{-0.5em}
\end{table}

Fig.~\ref{fig:res1} shows the comparison results of the classific Top-K algorithm and our proposed AdapTop-K on the MNIST dataset. Fig.~\ref{res:a} and Fig.~\ref{res:b} show the test accuracy curves and the training loss curves on the MNIST dataset. It shows how the model performance changes with iterations for several different values of the sparsification factor (128 or 256). The accuracy of the original distributed SGD reaches 98.02\%.  In Fig.~\ref{res:a}, the AdapTop-K achieves 97.03\% accuracy which is better than 96.64\% from Top-K. In Fig.~\ref{res:b}, the AdapTop-K achieves 96.21\% accuracy which is higher than 95.41\% from Top-K. The curve corresponding to the AdapTop-K achieves better performance than fixed Top-K compression when the compression ratios ($\frac{d}{k}$) are 128 and 256, respectively. 

Similarly, Fig.~\ref{fig:res2} shows the comparison results of the fixed Top-K and our proposed AdapTop-K on CIFAR-10 dataset. Fig.~\ref{res:d}
and Fig.~\ref{res:e} show the test accuracy curves and the training loss curves. It shows how the model performance changes with iterations for several different values of the sparsification factor (128 or 256). The accuracy of the original distributed SGD reaches 90.92\%. In Fig.~\ref{res:d}, the AdapTop-K achieves 82.11\% accuracy which is better than 81.36\% from Top-K. In Fig.~\ref{res:e}, the AdapTop-K achieves 80.31\% accuracy which is higher than 79.30\% from Top-K. The curve corresponding to the AdapTop-K achieves better performance than fixed Top-K compression when the compression ratios ($\frac{d}{k}$) are 128 and 256, respectively. 
We keep the communication cost of the AdapTop-K stable compared with the classic Top-K in the total training process. It can be seen that our adaptive sparsification strategy can effectively improve the convergence rate and model performance with the pure Top-K algorithm.
Fig.~\ref{res:c} and Fig.~\ref{res:f} both show the gradient sparsification level in the training process of  AdapTop-K on different datasets. We can see that AdapTop-K significantly increases the bits assigned at the early stage and the late stage of training and improves the gradient accuracy as the training goes on.

After that, we add the error compensation~\cite{wu2018error} (abbreviated as ec) in Fig.~\ref{fig:res3} and Fig.~\ref{fig:res4} in our experiments, because it is a popular technique to improve the performance of distributed SGD with gradient compression. It shows how the model performance changes with iterations for several different values of the sparsification factor (256 or 512) when we add the error compensation. In these experiments, we use the bigger compression ratios (e.g., 256 and 512) because error compensation may reduce optimization errors in the training process to improve the total performance. Fig.~\ref{fig:res3} and Fig.~\ref{fig:res4} show the comparison results of the classific Top-K algorithm and our proposed AdapTop-K (all with error compensation) on MNIST and CIFAR-20 datasets. In Fig.~\ref{res3:a}, the AdapTop-K achieves 97.50\% accuracy which is higher than 96.71\% from Top-K. In Fig.~\ref{res3:b}, the AdapTop-K achieves 97.10\% accuracy which is better than 96.24\% from Top-K. In Fig.~\ref{res4:a}, the AdapTop-K achieves 89.18\% accuracy which is better than 88.66\% from Top-K. In Fig.~\ref{res4:b}, the AdapTop-K achieves 88.68\% accuracy which is higher than 87.64\% from Top-K. The curve corresponding to the AdapTop-K achieves better performance than fixed Top-K compression when the compression ratios ($\frac{d}{k}$) are 256 and 512, respectively. The results show that the AdapTop-K algorithm with error compensation achieves better performance under stable communication cost. 
Overall, the evaluation results demonstrate that the AdapTop-K outperforms the baselines.

\section{Conclusion}
This paper proposes AdapTop-K, a novel adaptive gradient sparsification
strategy for distributed SGD. The proposed method adjusts the sparsification levels adaptively by considering the gradient and the current iteration step. The experimental results for image classification show that AdapTop-K is superior to the state-of-the-art gradient compression methods in reducing the communication cost.
\vspace{-0.05in}
\section*{Acknowledgment}
The work described in this paper was substantially sponsored by the project 62101471 supported by NSFC and was partially supported by the Shenzhen Research Institute, City University of Hong Kong. The work was also partially supported by the Research Grants Council of the Hong Kong Special Administrative Region, China (Project No. CityU 21201420 and CityU 11201422), Shenzhen Science and Technology Funding Fundamental Research Program (Project No. 2021Szvup126), Shandong Provincial Natural Science Foundation (Project No. ZR2021LZH010). This work was supported in part by the CityU grants 7005660, 7005849, InnoHK initiative, the Government of the HKSAR, Laboratory for AI-Powered Financial Technologies.

\vspace{-0.05in}
\section{Appendix}
\subsection{Proof for Theorem 1}
\vspace{-0.05in}
Using Eq.~\eqref{3.1} and Assumption~\ref{ass:1}, we get:
\vspace{-0.1in}
\begin{equation}
\begin{aligned}
&\E[F(\w_{t+1})]\leq F(\w_t)-\eta\langle\nabla F(\w_t), \C(\g_t)\rangle+\frac{\eta^2 L}{2}\E\|\C(\g_t)\|^2\\
&(\text{use} \quad \E\|\C(\g_t)\|^2= \E\|\C(\g_t)-[\C(\g_t)-(\g_t-\nabla F(\w_t))]\|^2+\\
&\quad\E\|\E[\C(\g_t)-(\g_t-\nabla F(\w_t))]\|^2\text{and Assumption~\ref{ass:3}}) \\
&\leq F(\w_t)-\eta\langle\nabla F(\w_t), \E[\C(\g_t)-(\g_t-\nabla F(\w_t))]\rangle\\
&\qquad+\frac{\eta^2 L}{2}( \sigma^2+\E\|\C(\g_t)-(\g_t-\nabla F(\w_t))\|^2) \\
&\leq F(\w_t)+\frac{\eta}{2}(\E\|\C(\g_t)-(\g_t-\nabla F(\w_t))\|^2)\\
&\;-2\langle\nabla F(\w_t), \E[\C(\g_t)-(\g_t-\nabla F(\w_t))]\rangle)+\frac{\eta^2 L}{2}\sigma^2\; (\eta\le\frac{1}{L}) \\
&\text{(from} \qquad \E\|\nabla F(\w_t)+\C(\g_t)-\g_t\|^2=\E\|\nabla F(\w_t)\|^2+\\
&\qquad\E\|\C(\g_t)-\g_t\|^2+2\E\langle\nabla F(\w_t), \C(\g_t)-\g_t)\rangle)\\
&\leq F(\w_t)+\frac{\eta}{2}(\E\|\C(\g_t)-\g_t\|^2-\E\|\nabla F(\w_t)\|^2)+\frac{\eta^2 L}{2}\sigma^2\quad\\
&(\text{from} \quad Eq.~\eqref{3.4} \text{ and assume that $k_t=k+n_t$, we have:} \\
&\quad\E\|b_t(\w)\|^2=\E\|\g_t-\C(\g_t)\|^2\leq\E[(1-\frac{k_t}d)\|\g_t\|^2]\\
&\qquad\leq\E[(1-\frac{k}d)\|\nabla F(\w_t)\|^2+(1-\frac{k}d)\sigma^2-\frac{n_t}{d}\Vert \g_t\Vert^2 ],\\
&\qquad\text{then put this equation back to our above derivation)} \\
&\leq F(\w_t)-\frac{\eta k}{2d}\|\nabla F(\w_t)\|^2+\frac{\eta }{2}(1-\frac{k}{d}+\eta L)\sigma^2-\frac{\eta n_t}{2d}\Vert \g_t\Vert^2.
\nonumber
\label{E.1}
\end{aligned}
\end{equation}
Therefore, we use Assumption~\ref{ass:2} and get convergence rate as
\begin{equation}
\begin{aligned}
\E[F(\w_{t+1})]-F^*&\leq (1-\frac{\eta k\mu }{d})(\E(F(\w_t)-F^*)\\
+\frac{\eta }{2}&(1-\frac{k}{d}+\eta L)\sigma^2-\frac{\eta n_t}{2d}\Vert \g_t\Vert^2.
\nonumber
\end{aligned}
\end{equation}

After recursion and simplification, we get:

\begin{equation}
\begin{aligned}
\E[F(\w_{T})]-F^*&\leq (1-\frac{\eta \mu }{d}k)^T[\E [F(\w_0)]-F^*]\\
&+\frac{d }{2k\mu}(1-\frac{k}{d}+\eta L)\sigma^2[1-(1-\frac{\eta \mu }{d}k)^{T}]\\
&-\sum_{t=0}^{T-1}[(\frac{\eta n_t}{2d}\Vert \g_t\Vert^2)(1-\frac{\eta \mu }{d}k)^{T-1-t}].
\nonumber
\end{aligned}
\end{equation}


\end{document}